# An Integrated Image Filter for Enhancing Change Detection Results

Dawei Li[1,2], Siyuan Yan[1], Xin Cai[1,2], Yan Cao[1], and Sifan Wang[1]
1. College of Information Sciences and Technology, Donghua University, Shanghai, China 201620
2. Engineering Research Center of Digitized Textile & Fashion Technology, Ministry of Education, Donghua University, Shanghai, China 201620
daweili@dhu.edu.cn(D. Li); xcai@dhu.edu.cn(Xin Cai)

## Abstract

*Change detection is a fundamental task in computer vision. Despite significant advances have been made, most of the change detection methods fail to work well in challenging scenes due to ubiquitous noise and interferences. Nowadays, post-processing methods (e.g. MRF, and CRF) aiming to enhance the binary change detection results still fall short of the requirements on universality for distinctive scenes, applicability for different types of detection methods, accuracy, and real-time performance. Inspired by the nature of image filtering, which separates noise from pixel observations and recovers the real structure of patches, we consider utilizing image filters to enhance the detection masks. In this paper, we present an integrated filter which comprises a weighted local guided image filter and a weighted spatiotemporal tree filter. The spatiotemporal tree filter leverages the global spatiotemporal information of adjacent video frames and meanwhile the guided filter carries out local window filtering of pixels, for enhancing the coarse change detection masks. The main contributions are three: (i) the proposed filter can make full use of the information of the same object in consecutive frames to improve its current detection mask by computations on a spatiotemporal minimum spanning tree; (ii) the integrated filter possesses both advantages of local filtering and global filtering; it not only has good edge-preserving property but also can handle heavily textured and colorful foreground regions; and (iii) Unlike some popular enhancement methods (MRF, and CRF) that require either a priori background probabilities or a posteriori foreground probabilities for every pixel to improve the coarse detection masks, our method is a versatile enhancement filter that can be applied after many different types of change detection methods, and is particularly suitable for video sequences. Experiments demonstrate that our method are suitable to be applied after a wide range of change detection methods, and it also works effectively on various scenes. In comparison, it evidently outperforms seven popular enhancement approaches.*

## 1. Introduction

Change detection aims at segmenting moving objects from images or throughout a video sequence. It is a fundamental task in computer vision, and now serves in important applications including intelligent security, national defense, intelligent transportation, etc. Change detection has been a popular research topic of the machine vision field for several decades.

Since the rise of modern computer vision research, a considerable amount of work has been done on change detection methodologies to adapt themselves to challenging and complex scenes. The classical change detection algorithms include Frame differencing [1], optical flow [2] and Gaussian Mixture Models (GMM) [3], [4]. Since the Millennium, substantial research achievements have been emerging rapidly, e.g., Kernel Density Estimation (KDE) based foreground detection [5], unsupervised clustering-inspired codebook algorithm [6], foreground detection based on random strategies [7], the robust principal component analysis (RPCA) approach in which the background is modeled by a low-rank subspace and the foreground components are regarded as a noise component [8-10], and video object segmentation via deep learning [11-13].

Despite significant advances have been made, rarely an algorithm is able to handle a enough broad scope of challenging scenes due to the ubiquitous interferences such as dynamic background, varying illumination, camera jitter, and shadow disturbances. Therefore, researchers resorted to post-processing methods to enhance the detection results and to generate satisfactory masks. The objective of the change detection enhancing algorithms is to reduce the detection noise and to make the foreground contours more accurate at the same time.

Traditional post-processing methods can date back to the morphologic operators [14, 15]. Although they are fast and easy to implement, the enhancement effect is usually unstable and far from satisfaction. Then, Markov Random Field (MRF) was widely used in post-processing for enhancing change detection results. In [16], Jodoin et al.

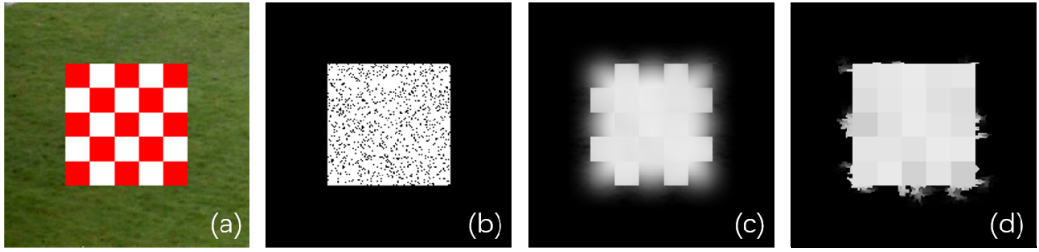

Figure 1: Comparison of local and global image filtering for enhancing a binary mask from a simulated change detection task. Fig. 1(a) is the original image (also treated as the guidance image), in which the foreground is a chessboard-like square object, and the background is the grass area that surrounds it; (b) is the coarse detection result after change detection; it is also treated as the input of filtering; (c) is the local guided filtering result on (b) with parameters setting at $r=10$, and $\varepsilon = 0.1 \times 0.1$. (d) is the global tree filtering result on (b) with $\sigma = 5$. The global filter has better performance over the local filter for this example.

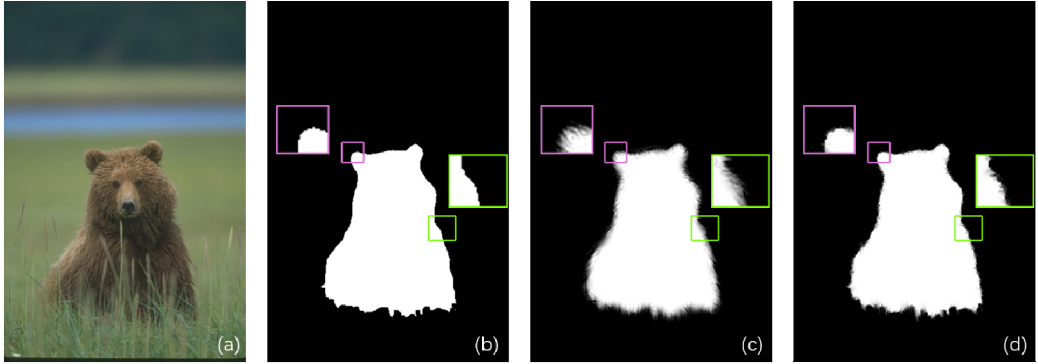

Figure 2: Comparison of local and global image filtering for enhancing a binary saliency detection mask. Fig. 2(a) is the original image (also treated as the guidance image), in which the foreground is a bear, and the background is the natural scene around it; (b) is a coarse result generated by the saliency detection algorithm in [33], and (b) is also the input of filtering; (c) is the local guided filtering result on (b) with parameters at $r=4$, $\varepsilon = 0.01 \times 0.01$. (d) is the global tree filtering result on (b) with $\sigma = 5$. The local filter has better performance over the global filter for this case because the former has better edge preserving ability than the latter.

apply a MRF model to change labels in binary detection masks by considering spatial coherence of the pixels. The model is in line with the argument that the probability of a pixel belonging to foreground is related to the number of foreground pixels as well as the number of background pixels in a surrounding area, and the accuracy of the foreground detection can be improved step by step in iterations of posterior probability computations. Schick et al. [17] improve the foreground detection with probabilistic superpixel MRF. Li et al. [18] think the prior probability of a pixel as the foreground is related to its neighborhood, and by defining a global energy function, they devise a pixel-wise MRF model iteratively computes the posterior probabilities belonging to the foreground and the background, respectively. Before 2012, only local-range conditional random fields (CRF) are used as post-processing to enhance the results of semantic image segmentation due to the formidable computation burden [19-21]. In 2012, [22] presents an efficient mean field approximation inference algorithm for fully-connected CRF models. Since then, dense CRFs became popular in the post-processing of deep learning based segmentation for their accurate boundary recovery ability and reliable enhancement performances [23], [24]. Though the approximated dense CRFs greatly reduce the computational complexity, they are still unfriendly to real-time applications.

Recently, advanced image filtering has become a hotspot of the image processing research. It is widely used in image denoising [25], [26], image haze removal [27], stereo matching [28], etc. Image filtering is to separate noise from pixel observations and to recover the real structure of pixels. The result of change detection is usually a coarse binary image, which can be conveniently enhanced by advanced image filters. A category of state-of-the-art filters [27-32] requires a guidance image related to the input image during filtering. They regard the information of guidance image as a reliable direction for filtering the input image. In the process of change detection, a detection mask is generated from a frame of a video sequence; so the pair of a video frame and its corresponding binary detection mask is information-correlated. Then it is natural to treat the original video frame as the guidance image when the detection mask needs to be improved. By utilizing the rich color and gradient information of the guidance image, a better mask can be obtained after filtering.

Image filtering can be divided into local image filtering and global image filtering according to whether the filtering process needs to be conducted in a moving window. Though local image filters (such as guided image filtering [27]) have the ability of capturing fine edge details, they fail to cater for long range dependency of pixels. Therefore, it has poor performance on objects with rich textures and colors. In a global filter, every pixel in the image can influence all the other pixels during filtering, and this global pixel coherence well handles the rich texture problem in foreground. However, it is not as good as local filtering regarding to the edge-preserving property. By comparing the enhancement effects of the representative of local filters—guided image filtering and the representative of global image filters—tree filter [28] on two change detection examples, we demonstrate the strength and weakness of the two types of the image filters, respectively. Fig. 1 shows the superiority of global filtering in handling textured foreground regions. Fig. 1(a) is a guidance image in which a simulated chessboard foreground region covers a patch of the lawn background. Fig. 1(b) is a noisy detection mask generated by a simple change detection procedure, and it is also the input image in filtering. In Fig. 1(c), it is obvious that the guided filter performs poorly around the boundaries between the white patch and the red patch in the chessboard like region. Moreover, the closer to the edge of the foreground, the darker the foreground pixel value becomes. Fig. 1(d) is the result of tree filtering. Tree filter considers the coherence of all pixels in the image so the pixels of the same color at a long distance can effectively support each other. After the enhancement of the global filter, the textured foreground presents a uniform white color and the scattered background noise in Fig. 1(b) disappear. Fig. 2 shows the advantage of local filtering in preserving foreground edges. Fig. 2(a) is a bear image as the guidance image. Fig. 2(b) is a rough detection image generated by a saliency detection algorithm proposed in [30]. Though the algorithm successfully sketches the general shape of the bear, it loses the hairy detail near boundary. The local filter has far better edge recovery result than the global tree filter in this case. It can be observed that the enlarged areas in Fig. 2(c) have richer information than their counterparts in Fig. 2(b) and Fig. 2(d), respectively. Therefore, the motivation of this research is to design a new image filter that can combine both advantages of local filtering and global filtering to maximize the enhancement effect on change detection results.

In this paper, we present an integrated filter which comprises a weighted local guided image filter and a weighted spatiotemporal tree filter for enhancing the coarse change detection masks. The spatiotemporal tree filter leverages the global spatiotemporal information of adjacent video frames and meanwhile the guided filter carries out local window filtering of pixels. The three main contributions are:

(i) Currently most of the non-local image filtering methods (e.g. CRF, and non-local tree filtering) only focus on the non-local spatial coherence of pixels in one image, ignoring the spatiotemporal correlation in image sequences. To bring in the potential similarity among images of the same foreground target in different epochs, we propose a global spatiotemporal tree filtering framework to effectively utilize the temporal relations between adjacent video frames. This filter can make full use of the information of the same object in several frames to improve its current detection mask by allowing all pixel support weights to transmit along paths on the minimum spanning tree. Therefore, the proposed spatiotemporal tree filtering extends the original tree filter from the spatial domain to the time domain. By introducing the historical features of the foreground objects, the spatiotemporal tree filtering is well suited to detect missing foregrounds and is also robust against noise.

(ii) We propose an integrated filter based on weighted global spatiotemporal filtering and weighted local guided filtering. The integrated filtering possesses both advantages of local and global filtering; it not only has good edge-preserving property but also can handle heavily textured and colorful foreground regions well. It is proved to be superior to popular state-of-the-art enhancement algorithms in both qualitative and quantitative experiments.

(iii) Currently, some popular change detection enhancement methods (e.g., MRF, and CRF) require either a priori background probabilities or a posterior foreground probabilities for every pixel to improve the coarse detection masks, which restricts their applications. Our method only needs several consecutive history frames and their corresponding detection masks to enhance the change detection mask in the current time. So our method is a versatile enhancement filter that can be applied after many different types of change detection methods, and is particularly applicable to video sequences.

The remainder of this paper is organized as follows. Section II first reviews related methodologies. Then, the global spatiotemporal tree filtering for image sequences is specified, followed by an elaboration of the integrated filtering based on a weighted sum of the local guided filter and the global spatiotemporal tree filter. At the end of Section II, a detailed workflow of the change detection enhancement framework based on our integrated filtering is explained. Experimental results and analysis are given in Section III and conclusions are drawn in Section IV.

2. Methods

A new type of image filters (e.g. joint bilateral filter [31], guided filter [27], and non-local tree filter [28]) utilize a guidance image relevant to the input image to enhance the detection mask (input image) during filtering. In change detection, the foreground mask, which usually takes the form of a binary image, is obtained from a video frame by applying the detection algorithm. The resultant masks are usually rifle with incorrect pixels such as false positives and false negatives. The original video frame has rich color, texture, and gradient information that the foreground mask does not possess; thus it is natural to regard the original video frame as the guidance image to improve the coarse foreground mask in filtering.

Both the joint bilateral filter and the guided filter are local methods that can enhance the input image by virtue of the guidance image information. The most attractive feature of the two local filters is the edge-preserving ability. In terms of the filtering effects, each has its own merits; but for presenting image details, the guided filter performs slightly better than the bilateral filter. In addition, compared with joint bilateral filtering, the complexity of guided filtering is independent of the size of filtering window. Therefore, when dealing with a large batch of images, guided filtering seems to be more efficient than the bilateral filtering. The non-local tree filter can boil down to a normalized linear image filter. Unlike local filters, tree filtering is a global spatial filtering which considers coherence among all pixels by allowing support information aggregates along paths on a minimum spanning tree. It uses a concept that the support between two distant pixels should flow in a way that costs least. However, the tree filter does not consider pixel relations in the time domain in filtering.

In this section, we intend to make use of the advantages of both local guided filtering and tree filtering to design an integrated image filter to enhance coarse change detection results. Hence, in this section we first separately introduce some basics of the two filters mentioned above. Then we describe the spatiotemporal tree filtering that combines the spatial pixel coherence and the temporal relations of video sequence. After that, an integrated filter combining a weighted global spatiotemporal filter and a weighted local guided filter is proposed to. Finally, a change detection mask enhancement algorithm with the proposed integrated Filtering is presented in the last subsection.

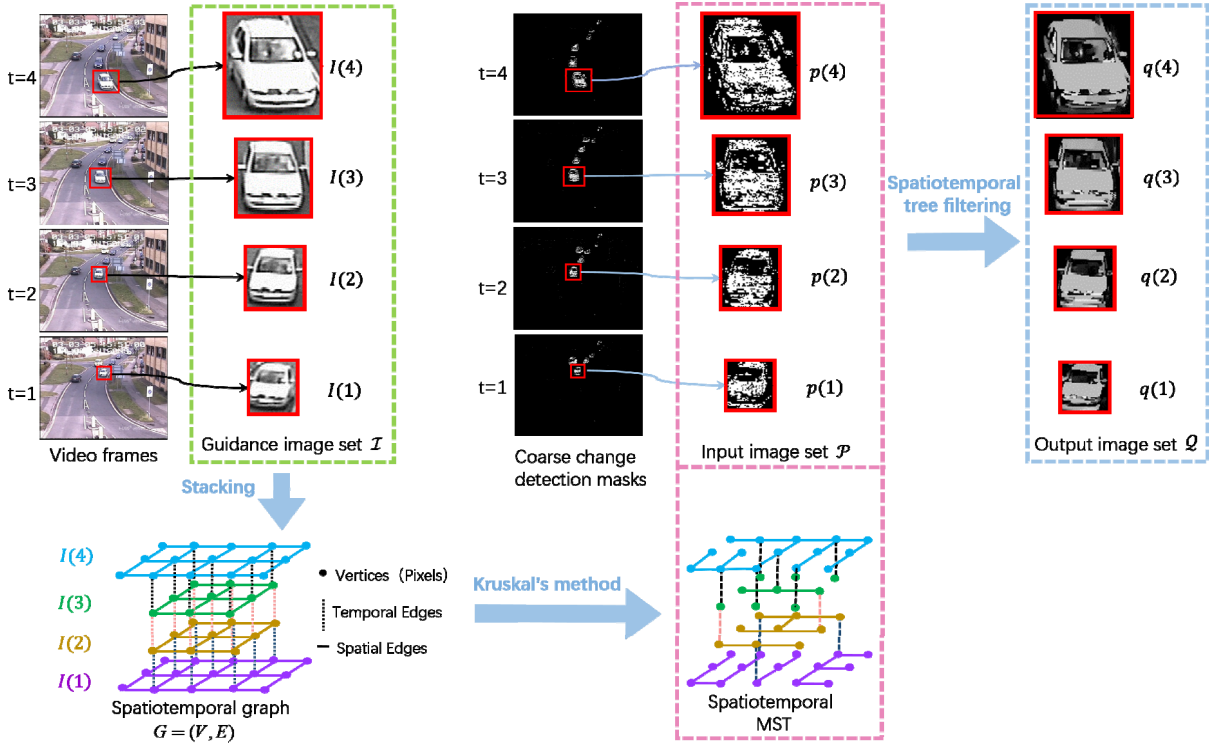

Figure 3: Pipeline of the proposed spatiotemporal tree filtering with $k=4$. Firstly, a spatiotemporal graph is formed by stacking the guidance images of the same object from several consecutive frames, respectively. Secondly, Kruskal's method is utilized to calculate the minimum spanning tree (MST) of the spatiotemporal graph. Lastly, an input image set that corresponds to the guidance image set is formed from coarse change detection results, and then the tree filtering is carried out on the input by exploiting the spatiotemporal MST. The enhanced output images are grayscale images shown in the rightmost part of the diagram.

2.1. Guided image filtering

Guided image filtering, a local linear filtering process in a window area, utilizes the rich color and orientation content in a guidance image to improve the input image by removing noise and enhancing foreground at the same time. It presumes that for the input image $p$ and the output image $q$ there exists a model formulated as $q = p - n$, in which $n$ can be regarded as some unwanted components like noise. Furthermore, it also hypothesizes that the output image $q$ has a linear relationship with the guidance image $I$ to emphasize the edge-preserving ability. The degree of influence of $I$ exerted on $q$ is controlled by the parameter $\varepsilon$. So guided filtering can be directly applied to enhance the result of change detection. For a standard guided filter, the original color image is first converted to a grayscale frame that is further used as the guidance image $I$ to improve a coarse binary mask $p$ obtained by a change detection algorithm; then a grayscale output image $q$ is generated after filtering. The filtering output $q_i$ at a pixel $i$ is

$$q_i = \sum_j \omega_{ij}(I) p_j, \quad (1)$$

in which $\omega_{ij}(I)$ is the weight between pixel $j$ and $i$. $j$ represents the pixel index in the window $W$ and $i$ is the center of the window.

$$\omega_{ij}(I) = \frac{1}{|W|^2} \sum_{(i,j) \in W} \left[ 1 + \frac{(I_i - \mu_W)(I_j - \mu_W)}{\sigma_W^2 + \varepsilon} \right], \quad (2)$$

$W$ is a square area with a radius of $r$. $\mu_W$, $\sigma_W$ denote the mean and variance of guidance $I$ in window $W$, respectively. $|W|$ is the pixel number in $W$. The filtering process is controlled by two parameters: the regularization parameter $\varepsilon$ and the window radius $r$. The output of guided filtering is a grayscale image, in which the whiter part indicates the higher likelihood of belonging to foreground, and the darker part is more likely to be background.

2.2. Tree filtering

Tree filter is a global image filter proposed by Yang *et al.* in [28]. For any pixel in the input image, it is supported by all pixels in the guidance image by receiving weights through the minimum spanning tree (MST). Similar to guided filtering, tree filtering regards the grayscale frame as the guidance image $I$. The guidance image $I$ is treated as a 4-connected, undirected graph $G = (V, E)$, in which the vertices are all the pixels in the guidance image and the edges are 4-connection relations between neighboring pixels. The weight of an edge connecting two nodes (pixels) $u$ and $v$ is decided by the absolute value of the intensity difference between $I(u)$ and $I(v)$:

$$e(u,v) = |I_u - I_v|. \quad (3)$$

It is straightforward that $e(u,v) = e(v,u)$. A minimum spanning tree (MST), which connects all nodes without forming circles and has the smallest sum of the weights in all spanning trees, is generated from the graph $G$ defined above. The similarity $S_I$ between any two nodes $u$, $v$ is defined as

$$S_I(u,v) = S_I(v,u) = \exp\left(-\frac{L_I(u,v)}{\sigma}\right), \quad (4)$$

where $L_I(u,v)$ is the length between the two nodes, calculated as the summed weights between two nodes $u$, $v$ on the MST. Let the original binary detection mask as the input image, the tree filtering output $q_i$ for pixel $i$ is:

$$q_i = \frac{\sum_j S_I(i,j) p_j}{\sum_j S_I(i,j)} = \frac{\sum_j \exp\left(-\frac{L_I(i,j)}{\sigma}\right) p_j}{\sum_j \exp\left(-\frac{L_I(i,j)}{\sigma}\right)}, \quad (5)$$

in which, $j$ represents all the pixel index in the image, including $i$ itself. $\sigma$ controls the degree of difficulty for information passing on the MST. Akin to guided filtering, the output $q$ of a tree filter is also a grayscale image, in which a blighter pixel is more likely to be classified to the foreground.

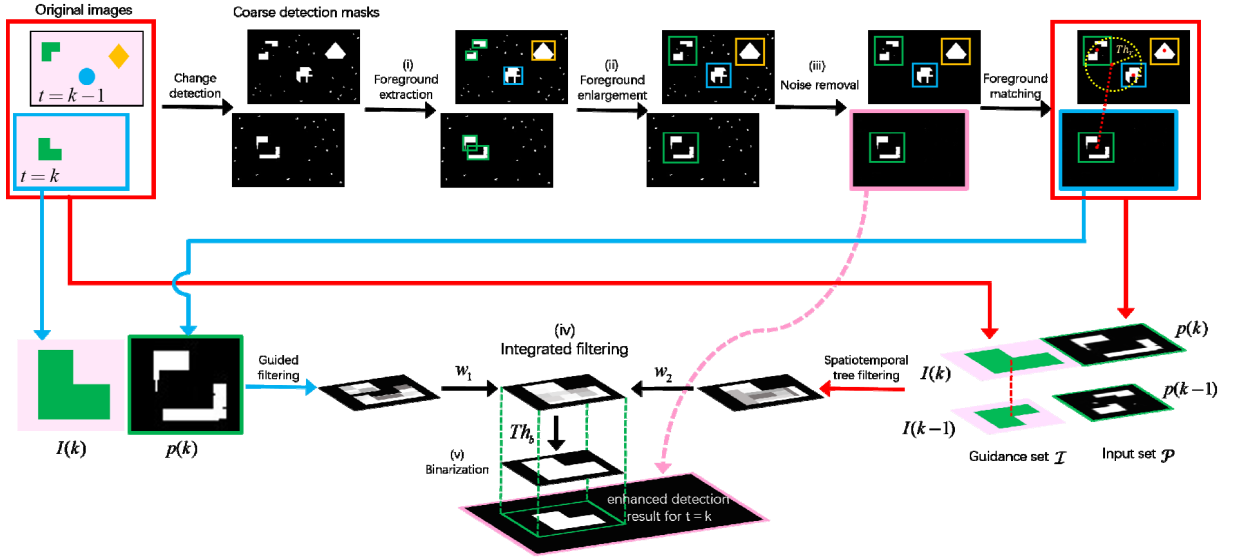

Figure 4: A workflow of the proposed integrated filtering for change detection result enhancement. The proposed algorithm contains 5 steps: (i) foreground extraction, (ii) foreground enlargement, (iii) noise removal, (iv) integrated filtering, and (v) binarization. The guided filtering part in the integrated filtering is highlighted in blue arrows, and the spatiotemporal tree filtering part is underscored by red arrows.

2.3. Spatiotemporal tree filtering

The video sequences have the characteristics of continuity in time and during a period of time a moving object often exhibits posture changes, self-rotation, color changes (e.g., stepping into a shadow). Due to the differences in the background-foreground color contrast at different times, the detection results of the same object can be very different. In addition, due to the noise and the changes in object posture, regions with new color and texture may probably appear on the same object. The image information in a spatiotemporal volume is a promising information source to create better detection results because the movement of an object is usually well kept in a period of time, and the successive images of the same object are intrinsically similar. Therefore, we consider the spatial and temporal aspects of foregrounds at the same time by constructing a spatiotemporal tree filtering framework to enhance the foreground detection results at the current time.

Figure. 3 shows the pipeline of the proposed spatiotemporal tree filtering process. We extract areas in bounding boxes of the same object at $k$ consecutive video frames to form a guidance image set $\mathcal{I}=\{I(t) | t \in \{1,...,k\}\}$. The guidance image of this object at time $t$ can be denoted by $I(t)$. The areas in the coarse detection results that corresponds to each guidance images constitute an input image set $\mathcal{P} = \{p(t) | t \in \{1,...,k\}\}$. The value of $k$ should be confined in a proper range. $k$ cannot be too large, because a moving object generally only appears in a short period of time during a video sequence. In order to make full use of the temporal pixel information, $k$ cannot be too small. A 6-connected, undirected spatiotemporal graph $G = (V, E)$ is built by stacking all images in the guidance image set $\mathcal{I}$ together with their center pixels aligned. The newer a guidance image is, the higher it is stacked. The vertex set $V$ contains all pixels in the spatiotemporal volume. The connection between adjacent pixels from two different layers forms a temporal edge. The spatial edges are edges of the 4-connected grids in each guidance image. The weight of each edge in the spatiotemporal graph is defined by equation (3). When $k$ is small, the spatiotemporal graph is a sparse graph (the number of the nodes and the edges are of the same order of magnitude); thus Kruskal's method [34], which is efficient for a sparse graph, is utilized to calculate the MST of the spatiotemporal graph. The output at a pixel $i$ is filtered by the information transmitted from every pixel in the guidance image set by MST:

$$Q_i = \frac{\sum_j \exp\left(-\frac{L_\mathcal{I}(i,j)}{\sigma}\right) \mathcal{P}_j}{\sum_j \exp\left(-\frac{L_\mathcal{I}(i,j)}{\sigma}\right)}. \quad (6)$$

Equation (6) is analogous with (5) except that the information aggregation are in a spatiotemporal graph. A direct calculation of (6) has a complexity of $O(n^{5/2})$ (measured by CPU clock cycles), in which $n$ stands for the pixel number of $\mathcal{I}$. When the number of pixels in the guidance image is high, the computation cannot meet realtime requirement, Yang [28] proposed a linear-time algorithm for speeding up the calculation of tree filtering, which can be descended to compute the spatiotemporal filtering process of (6). The acceleration uses the relation between the root node and its child nodes on the MST to reduce repetitive computation. In this paper, the spatiotemporal tree filtering adopts this acceleration algorithm to compute the numerator and the denominator of (6) separately for assuring real-time enhancement of change detection results.

2.4. Integrated filtering

From the examples of Fig. 1 and Fig. 2, it can be clearly seen that the local guided filter is effective in edge-preserving filtering; but it fails to consider the pixel information beyond the window range. Although the global tree filtering considers the global pixel correlation, its edge-preserving effect is weaker than the guided filtering. Even though the spatiotemporal tree filtering adds foreground history information to the original non-local tree filter, the performance still has room for improvement. Therefore, we consider combining the benefit of global spatiotemporal tree filtering and the edge-preserving advantage of the local filtering to design a new integrated filter that is able to drive the detection mask to get closer to the ground truth:

$$q_i(k) = w_1 \frac{\sum_{j \in \mathcal{T}, i \in I(k)} S_{\mathcal{T}}(i,j) \mathcal{P}_j}{\sum_{j \in \mathcal{T}, i \in I(k)} S_{\mathcal{T}}(i,j)}$$
$$+ w_2 \sum_{\substack{i,j \in I(k) \\ (i,j) \in W}} \frac{1}{|W|^2} \sum_{\substack{i,j \in I(k) \\ (i,j) \in W}} \left[1 + \frac{(I_i(k) - \mu_W)(I_j(k) - \mu_W)}{\sigma_W^2 + \varepsilon}\right] p_i(k). \quad (7)$$

The integrated filter formulated by (7) have two terms in the right hand side. The first term stands for the spatiotemporal tree filtering process, and the second term is a standard guided filter. The weights assigned to the two parts must satisfy $w_1 + w_2 = 1$. The spatiotemporal tree filtering term utilizes cropped guidance images of a same moving object at $k$ consecutive times to enhance the coarse detection result at the current time. The guided filtering term makes use of the local spatial information confined in the current object image area to improve the edges of the foreground. In summary, the proposed integrated filtering combines dual-scale information—global scale and local scale, and it also integrates spatial information and temporal information at the same time. By adjusting the weights in filtering, we can control the ratio of different information types, which makes the integrated filtering to be highly flexible.

Since the results of the spatiotemporal tree filtering and the guided filtering are grayscale images, the result of the integrated filtering is also a grayscale image with all pixel values range between 0 and 255. Given a simple binarization threshold, the filtered mask can be immediately converted into a binary image. The discussion of fixing this binarization threshold will be given in the experimental section.

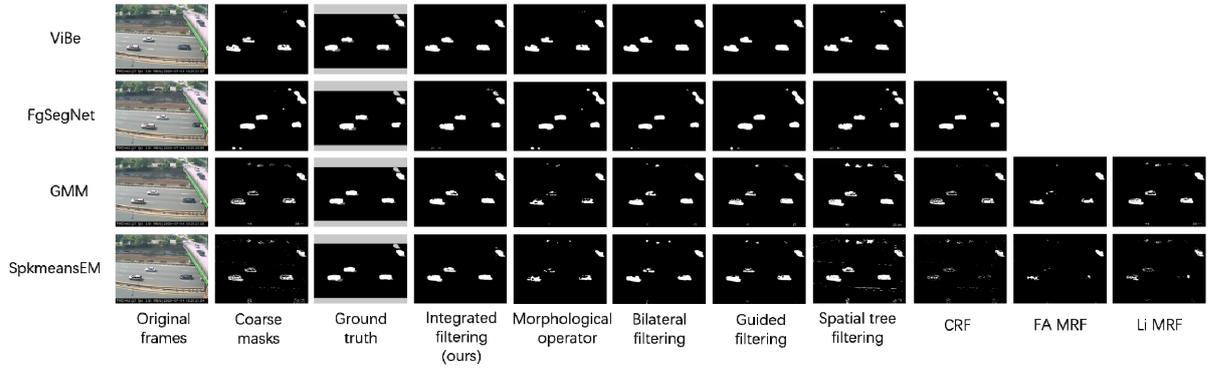

Figure 5: Qualitative comparison results on "streetlight" sequence. From the 1st row to the 4th row are results of ViBe, FgSegNet, GMM, and SpkmeansEM, respectively. In order to avoid visual redundancy, we use different frames for different change detection methods. From the 1st column to the 10th column are the original frames, binary masks after coarse detection, ground truth, results of our method, results of morphological operation, bilateral filtering results, guided filtering results, spatial tree filtering results, CRF results, FA MRF results, and Li MRF results. Our method outperforms other enhancement algorithms.

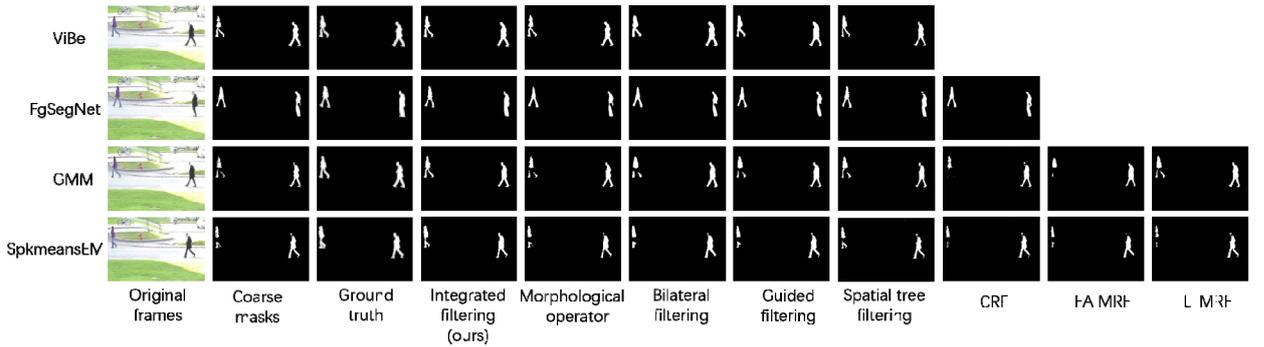

Figure 6: Qualitative comparison results on "pedestrians" sequence. From the 1st row to the 4th row are results of ViBe, FgSegNet, GMM, and SpkmeansEM, respectively. In order to avoid visual redundancy, we use different frames for different change detection methods. From the 1st column to the 10th column are the original frames, binary masks after coarse detection, ground truth, results of our method, results of morphological operation, bilateral filtering results, guided filtering results, spatial tree filtering results, CRF results, FA MRF results, and Li MRF results. Our method outperforms other enhancement algorithms.

## 2.5. Change detection mask enhancement with integrated filtering

In this section, we apply the proposed integrated filter to change detection mask enhancement. The framework of the entire algorithm can be divided into the following five steps: (i) foreground extraction, (ii) foreground enlargement, (iii) noise removal, (iv) integrated filtering, and (v) binarization. We specify in the following for the five steps. Fig. 4 illustrates the complete workflow of the algorithm.

(i) Foreground extraction: For coarse detection masks at $k$ consecutive frames obtained by a foreground detection method, we first carry out connected component labeling to extract potential foreground areas that have a number of white pixels higher than a threshold $Th_{area}$. Then we use a bounding box to circumscribe every potential foreground area. In the example shown in Fig. 4, after the foreground area extraction step, two green bounding boxes are generated in the mask at time $k$, while two green boxes, 1 blue box, and 1 yellow bounding box are generated in the mask at time $k-1$.

(ii) Foreground enlargement: Due to the limitation of change detection, some parts of foreground objects can be easily mis-classified to background and a single object may break into several isolated foreground regions in coarse masks. Therefore, we expand the size of the bounding box for each potential foreground by 10 to 20 pixels and merge the overlapped foreground boxes with a bigger detection box. After merging, potential foreground areas in $k$ frames are updated, which also means the number and the area of potential foregrounds can both change. In the example shown in Fig. 4, after foreground enlargement, only one green bounding box remains in the mask at frame $k$, and 3

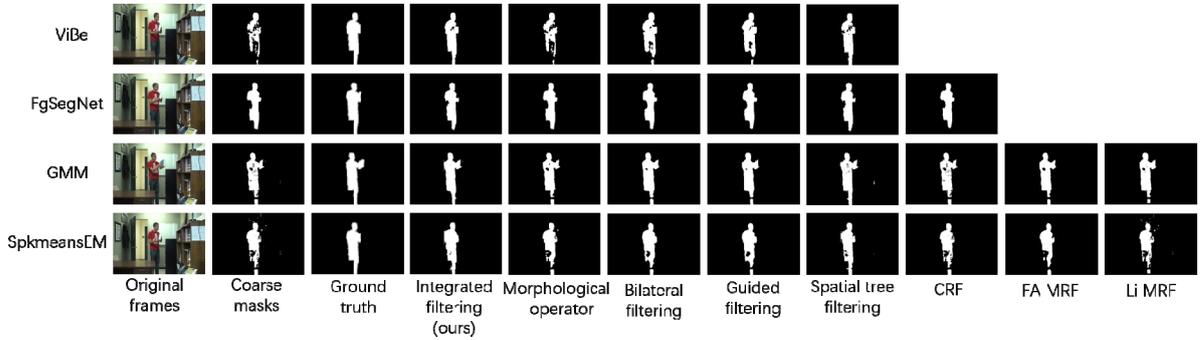

Figure 7: Qualitative comparison results on "office" sequence. From the 1st row to the 4th row are results of ViBe, FgSegNet, GMM, and SpkmeansEM, respectively. In order to avoid visual redundancy, we use different frames for different change detection methods. From the 1st column to the 10th column are the original frames, binary masks after coarse detection, ground truth, results of our method, results of morphological operation, bilateral filtering results, guided filtering results, spatial tree filtering results, CRF results, FA MRF results, and Li MRF results. Our method outperforms other enhancement algorithms.

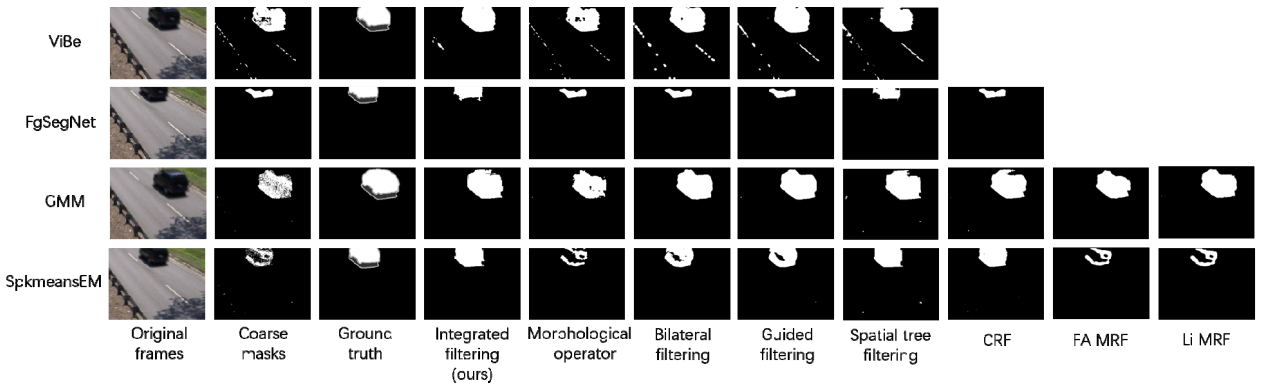

Figure 8: Qualitative comparison results on "traffic" sequence. From the 1st row to the 4th row are results of ViBe, FgSegNet, GMM, and SpkmeansEM, respectively. In order to avoid visual redundancy, we use different frames for different change detection methods. From the 1st column to the 10th column are the original frames, binary masks after coarse detection, ground truth, results of our method, results of morphological operation, bilateral filtering results, guided filtering results, spatial tree filtering results, CRF results, FA MRF results, and Li MRF results. Our method outperforms other enhancement algorithms.

boxes remains in the mask at frame $k-1$.

(iii) Noise removal: We assume that for the consecutive k-frame coarse detection masks obtained by a foreground detection algorithm, those pixels that are outside the merged potential foreground areas all belong to the background. So, dispersed white pixels outside the extended foreground regions are zeroed to realize noise removal.

(iv) Integrated filtering: Our integrated filtering can be decomposed into two main parts: guided filtering and spatiotemporal tree filtering. Guided filtering is performed at first. A foreground region in the coarse detection image at time $k$ is used as the input image and the corresponding region in the original image of the same video frame is used as the guidance image, the output is then multiplied by the weight $w_2$ to obtain the result of local filtering part. The result corresponds to the second part of the right-hand-side of equation (7). After guided filtering, the spatiotemporal tree filtering is performed. We design a foreground matching mechanism to build a reliable spatiotemporal volume for each foreground object. In this matching mechanism, the corresponding guidance region of each single potential foreground region at time $k$ is treated as the first image in the guidance set for this foreground object. Then, starting from the center of a foreground region at time $k$, we search in several past frames to see whether the distances between the original center at $k$th frame and the centers of other foreground regions at past are within the distance threshold $Th_r$. If so, it is considered that matching candidate for guidance images are found. If multiple matching candidates are found in one frame, we only keep the candidate that has the closest mean color to the original foreground region in frame $k$ to ensure that at each time only one target region can match. Finally the remaining matching candidates in all frames are stacked below the first guidance image to form a spatiotemporal volume, i.e., the guidance set. In the example shown in Fig. 4, two targets in frame $k-1$ fall into the circle whose radius is the distance threshold $Th_r$; however, only the target with green bounding box has a closer mean color intensity to the original foreground area in frame $k$. Note that if no matching target is found in the binary detection image at a certain time, the guidance set is unchanged and the search continues to an earlier frame. After retrieving all $k$ times, the guidance set turns into a spatiotemporal guidance volume, and the potential foreground regions of the target at all times are used as the input set to perform the spatiotemporal tree filtering. The output grayscale image at the current time $k$ is then multiplied by $w_1$ to acquire the result of the global spatiotemporal filtering part. The result is the first part of the right-hand-side of the (7).

An integrated filtering result, a grayscale image, is an addition of the local filtering result and the global filtering result.

TABLE 1
COMPARISON OF F-MEASURE VALUES ACROSS EIGHT ENHANCEMENT METHODS

| Sequences | Change detection methods | F-measure | | | | | | | | |
|---|---|---|---|---|---|---|---|---|---|---|
| | | Coarse masks | Enhancement methods | | | | | | | |
| | | | Morphological operator | FA MRF | Li MRF | Joint bilateral filtering | Guided filtering | CRF | Spatial tree filtering | Our method |
| Streetlight | GMM | 0.7199 | 0.6548 | 0.6585 | 0.8315 | 0.8463 | 0.8374 | 0.7024 | 0.8999 | 0.9264 |
| | SpcmeansEM | 0.6926 | 0.6495 | 0.4335 | 0.4978 | 0.8419 | 0.8442 | 0.4316 | 0.8301 | 0.9250 |
| | Vibe | 0.8958 | 0.9262 | \ | \ | 0.9252 | 0.9479 | \ | 0.9398 | 0.9589 |
| | FgSegNet | 0.9241 | 0.9245 | \ | \ | 0.9186 | 0.9276 | 0.9338 | 0.9275 | 0.9350 |
| Pedestrians | GMM | 0.9642 | 0.9702 | 0.9280 | 0.9895 | 0.9852 | 0.9867 | 0.9009 | 0.9877 | 0.9956 |
| | SpcmeansEM | 0.9455 | 0.9500 | 0.9295 | 0.9068 | 0.9734 | 0.9751 | 0.9010 | 0.9745 | 0.9945 |
| | Vibe | 0.9837 | 0.9876 | \ | \ | 0.9979 | 0.9887 | \ | 0.9895 | 0.9902 |
| | FgSegNet | 0.9703 | 0.9715 | \ | \ | 0.9721 | 0.9758 | 0.9791 | 0.9818 | 0.9894 |
| Office | GMM | 0.9391 | 0.9510 | 0.9563 | 0.9575 | 0.9586 | 0.9679 | 0.9401 | 0.9596 | 0.9697 |
| | SpcmeansEM | 0.8898 | 0.9015 | 0.8928 | 0.8664 | 0.9304 | 0.9305 | 0.8948 | 0.9500 | 0.9623 |
| | Vibe | 0.7896 | 0.8162 | \ | \ | 0.8827 | 0.8988 | \ | 0.9374 | 0.9617 |
| | FgSegNet | 0.9495 | 0.9501 | \ | \ | 0.9516 | 0.9526 | 0.9444 | 0.9578 | 0.9618 |
| Traffic | GMM | 0.8392 | 0.8937 | 0.8991 | 0.9332 | 0.9605 | 0.9639 | 0.9593 | 0.9663 | 0.9815 |
| | SpcmeansEM | 0.6441 | 0.6044 | 0.6254 | 0.6638 | 0.8979 | 0.9125 | 0.9759 | 0.9765 | 0.9835 |
| | Vibe | 0.9132 | 0.9369 | \ | \ | 0.9211 | 0.9286 | \ | 0.9520 | 0.9831 |
| | FgSegNet | 0.9270 | 0.9283 | \ | \ | 0.9325 | 0.9360 | 0.9364 | 0.9749 | 0.9896 |

The best results are in bold face, and the second best are underlined.

(v) Binarization: Since the filtering result is a grayscale image, a threshold $Th_b$ is used to binarize the filtered result for generating an enhanced foreground detection region. The enhance region then covers the corresponding region in the coarse detection mask after noise removal at time $k$ to obtain the final enhanced binary detection result. In Fig. 4, the output of the integrated filtering is binarized by the threshold $Th_b$, and the binarized foreground region with a green bounding box finally covers the corresponding area in the mask after noise removal step (with a pink boundary) to generate a final enhanced detection.

3. Experiments and analysis

In this section, we try to validate the superiority of the proposed image filter as well as its widespread applicability for enhancing many different kinds of change detection methods. In the subsection 3.1, we first conduct a qualitative assessment of our method and seven other methods by testing them on four video sequences from the CDnet2012 dataset [35]. The titles of the four sequences are the "streetlight", "pedestrians", "office", and "traffic". The adopted sequences are challenging, some are rifle with camera jitter, some contain similar foreground and background. For each sequence, we first use four foreground detection algorithms to obtain some consecutive coarse detection binary masks. The four foreground detection algorithms are GMM [4], ViBe [7], SpkmeansEM [18], FgSegNet [12]. GMM (Gaussian mixture model) is a classical statistical change detection algorithm based on the Gaussian mixture background model. SpkmeansEM (Spherical K-means Expectation Maximization algorithm) is an improved GMM background-foreground subtraction algorithm. ViBe is an efficient foreground detection method with random strategies. FgSegNet is a supervised deep learning algorithm based on Convolutional Neural Networks (CNN). The parameters of the four foreground detection algorithms are selected according to the recommendations in their respective papers. For qualitative comparison, the coarse detection masks from different change detection algorithms are improved by 7 enhancement algorithms at the same time; the 7 compared methods are the morphological operator [14-15], joint bilateral filtering [29], guided filtering [27], spatial tree filtering [28], fully-connected conditional random fields (CRF) [23], Foreground-adaptive MRF (FA MRF) [16], Li MRF [18], and the proposed integrated filter. In quantitative evaluation, we adopt *F-measure* as the test criterion:

$$F\text{-measure} = \frac{2 \times Precision \times Recall}{Precision + Recall}, \qquad (8)$$

in which *Precision* is defined as $TP/(FP+TP)$, and *Recall* is defined as $TP/(TP+FN)$. TP (True Positive) is the number of pixels that are both determined by the algorithm and ground truth as foreground. TN (True Negative) is the number of pixels that are both determined as background by the algorithm and the ground truth. FP (False Positive) is the number of pixels that are determined as foreground by the algorithm but are actually background. FN (False Negative) is the number of pixels that are determined as background by the algorithm but are foreground in ground truth.

The configuration of parameters in the integrated filtering is detailed in subsection 3.2. The morphological operation may be the earliest foreground enhancement algorithm. The morphological operator adopted in this paper contains an open operator and a close operator; the open operation runs first to remove isolated noise from mask and then the close operation runs to enhance the foreground boundaries. The parameter configurations for using guided image filtering alone and using it as a weighted part in our integrated filtering are the same: $r=4$, $\varepsilon=0.2\times 0.2$. The parameters of the joint bilateral filtering are set as: $\sigma_s=150$, $\sigma_r=0.2$, and $r=5$. The parameters of spatial tree filtering are the same as the parameters of the spatiotemporal tree filtering in the integrated filter. The outputs of guided filtering, joint bilateral filtering, and spatial tree filtering are all grayscale images, so a binarization threshold $Th_b$ is needed to convert the outputs into final binary images. For each of those methods, the threshold $Th_b$ is separately adjusted to generate the best performance for comparison. In realization of the CRF, we regard the binary detection mask as a 2-class segmented image and then use the Fully-Connected CRF adopted in DeepLab [23] to enhance the mask. The parameters of the CRF are tuned according to suggestions in [22]. We refer to the MRF enhancement method proposed by Jodoin *et al.* [16] as the foreground-adaptive MRF (FA MRF), and the MRF enhancement method proposed by Li *et al.* [18] is simply referred to as Li MRF. In all experiments, the number of iterations for MRF are no larger than 3 to prevent from foreground over-smoothing, while selection of other parameters is based on the values provided in original papers. Since ViBe is not a statistical change detection method, both CRF and MRF cannot be applied. Therefore, we only compare the proposed method, the morphological operation, joint bilateral filtering, guided filtering, and spatial tree filtering on ViBe. FgSegNet is not able to obtain the foreground prior probability for each pixel, thus it is not suited to MRFs. For FgSegNet, we only do comparison of the proposed method, the morphological operation, bilateral filtering, guided filtering, spatial tree filtering, and CRF.

The average quantitative results of the proposed algorithm for enhancing three other change detection algorithms on 12 complete sequences in CDnet2012 are given in subsection 3.3. The computational time costs of the eight methods compared are presented in subsection 3.4. All experiments are carried out on a PC operated by Windows 10 with a 3.4 GHz Intel Core i7 CPU and 16 GB memory.

3.1. Comparative results

1) Qualitative comparison

Figs. 5-8 visually compare the results of eight enhancement methods applied after four different change detection methods, respectively. The qualitative experiments are carried out on four video sequences of CDnet2012 dataset, with every sequence to be detected by all the four foreground detection algorithms. In Figs. 5-8, the rows from top to bottom are: comparison of five enhancement algorithms on the coarse detection mask obtained by ViBe, comparison of six enhancement algorithms on the results of FgSegNet, comparison of eight enhancement algorithms on the coarse masks of SGMM, comparison of the eight enhancement algorithms on coarse detection masks obtained by SpkmeansEM. In order to avoid redundancy, in a same sequence, we show results on 4 different change detection algorithms at 4 different frames, respectively. In Figs. 5-8, From the leftmost column to the rightmost column are the original frames, the coarse binary masks from change detection, ground truth, results of the proposed method, results of the morphological operation, bilateral filtering results, guided filtering results, spatial tree filtering results, CRF results, FA MRF results, and Li MRF results.

Fig. 5 is the comparison results on the "streetlight" video sequence. Except for FgSegNet, the other three detection methods miss a large part of the white car in the middle. For all four change detection methods, our algorithm greatly improves the contour integrity for all vehicles. Among the results on enhancing ViBe (the first row in Fig. 5), the proposed method is the closest to the ground truth. In the coarse mask of FgSegNet, we can see that the two cars in the top-right corner are over-smoothed, resulting in a false connection. Only our method, spatial tree filtering, and CRF can correctly separate the two cars. The result of the proposed integrated filter is similar to the spatial tree filter, but the former has less noise in the background area. On the third row, the coarse detection mask generated by GMM is not satisfactory. Expect for the car in the top-right corner, the detected foreground areas of other vehicles only account for less than half of the true areas. Joint bilateral filtering, guided filtering, In this case, the filter-based enhancement algorithms seem to be particularly effective, and among them, our method outperforms other enhancement results in completeness of foreground. As for the case of SpkmeansEM, only the joint bilateral filtering, guided filtering, and our method can enhance the coarse mask. By introducing the edge-preserving property of the guided filtering, our integrated filtering obtains accurately shaped foreground contours. Though the spatial tree filtering also obtains good result, it adds some additional noise after enhancement. MRFs consider the neighborhood information in computing the probability of foreground. Therefore, if the detected foreground pixel distribution is too sparse MRFs may even further dilute the already sparse foreground clusters. In the same sense, the unary term of CRF is not reliable enough when foreground pixels are sparse, which may easily deteriorates the coarse detection mask such as the example shown in the last row of Fig. 5.

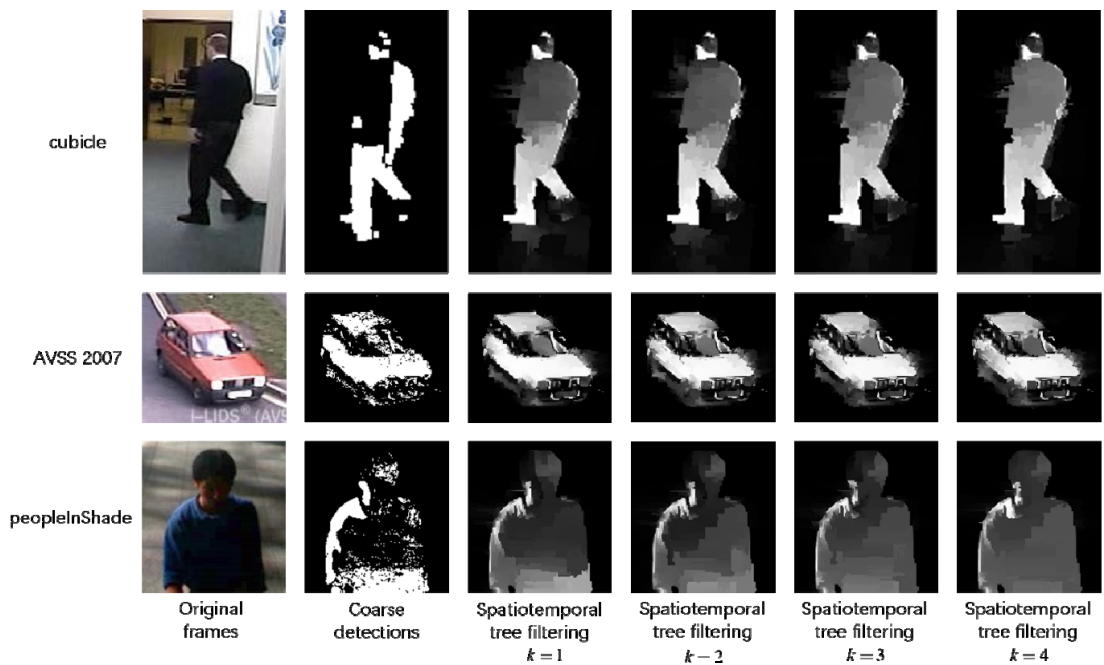

Figure 9: Influence of $k$ on the spatiotemporal tree filtering part of the integrated filter. Row 1 is the spatiotemporal filtering results on a foreground region from the "cubicle" sequence in the CDnet dataset. Row 2 shows the filtering results on a vehicle region from the AVSS2007 dataset. Row 3 shows the filtering results on a pedestrian region from the "peopleInShade" sequence in CDnet dataset. From left to right, the 6 columns are the original frames, the coarse detection masks by SpkmeansEM, the spatiotemporal filtering outputs at $k=1$, $k=2$, $k=3$, and $k=4$.

Fig. 6 presents the comparison results on the "pedestrians" sequence. In the coarse detection result of ViBe, the upper body of the left person is disconnected in the mask. All the five enhancement methods compared can re-connect the upper body fragment in the ViBe case, with the proposed integrated filter exhibiting the closest result to ground truth. The results of joint bilateral filtering and guided filtering are over-smoothed around the edges of two foreground objects, i.e., the foreground objects become larger than they really are. Under FgSegNet, a middle part of the right person's foreground is missing. Across all compared enhancement methods, our method is the only one that correctly fills the missing foreground area. The spatial tree filter also yields a good result; however, the head area of the right person is not fully detected. As for cases of GMM and SpkmeansEM, the thighs of the left person are missing in coarse masks; but our method successfully connects the disconnected foreground fragments and achieves the best enhancement performance across all 8 methods.

Fig. 7 shows the comparative results on the "office" sequence. Due to the reason that the color of the right leg of the target is almost the same as the color of the skirting line (background), a small foreground area in the middle of the right leg is not detected by Vibe, GMM, and SpkmeansEM. For the coarse mask of FgSegNet, our method can accurately fill the missing foreground part in the right leg

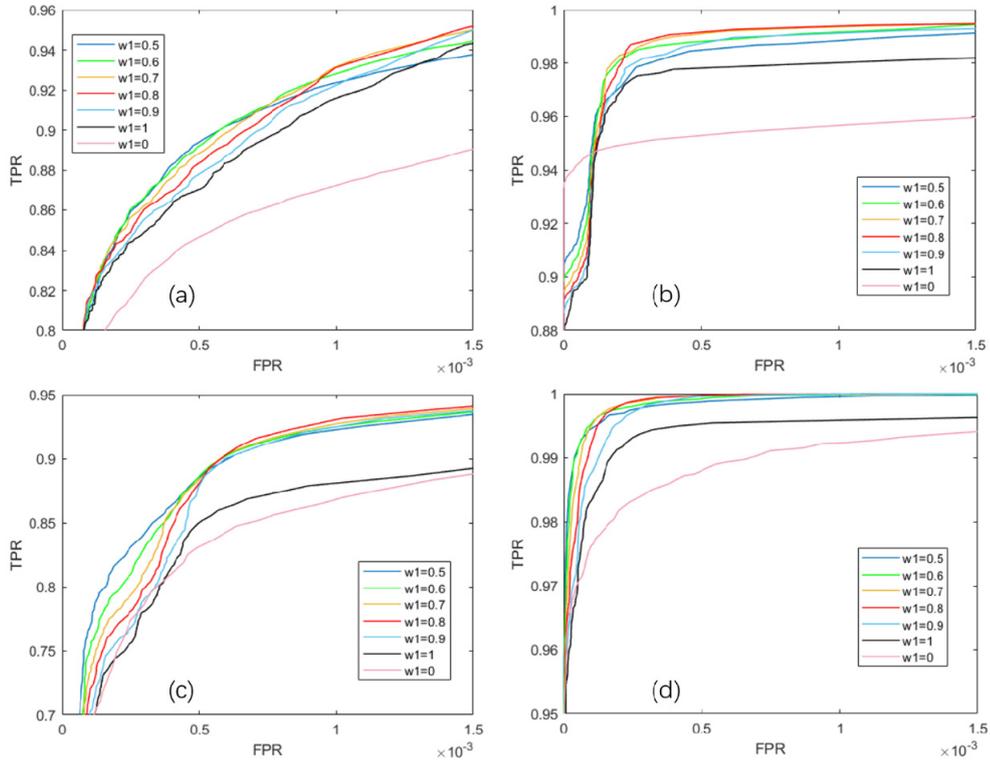

Figure 10: The comparison of ROC curves with different with different ratios between $w_1$ and $w_2$. Seven different ratios are compared, with $w_1$ set to 0, 0.5, 0.6, 0.7, 0.8, 0.9, and 1. (a) is the ROC comparison on "office" sequence with SpkmeansEM as the coarse change detection method; (b) is the ROC comparison on the "traffic" sequence with GMM as the coarse change detection method; (c) is the ROC comparison on "streetlight" sequence with GMM as the coarse change detection method; (d) is the ROC comparison on "pedestrians" sequence with SpkmeansEM as the coarse change detection method.

and also exhibits the best performance across all compared enhancement methods.

In Fig. 8, the FgSegNet algorithm only detects a small part of the vehicle. After enhanced by our method, the foreground rea obviously outcompetes other enhanced results, and is almost exactly the same as the ground truth. By observing the results on "traffic", it can be noted that the integrated filtering generates foreground areas that are highly intact and have more accurate contours.

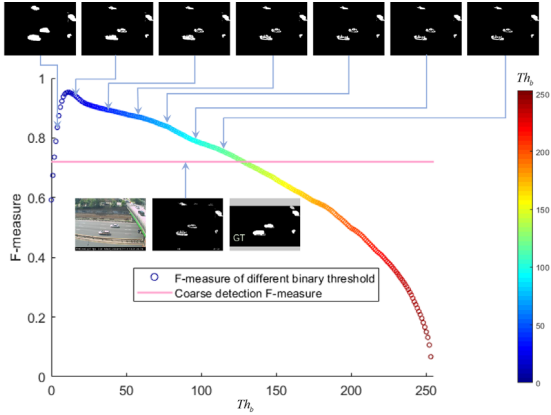

Figure 11: Qualitative and quantitative results with different values of $Th_b$ on a GMM detected frame from the "streetlight" sequence. We find $Th_b = 20$ obtains the optimal enhancement result for the proposed filter in this case.

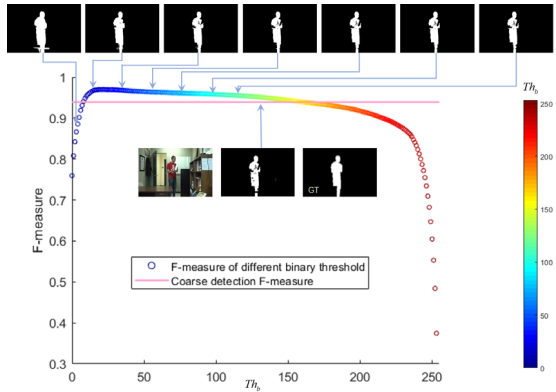

Figure 12: Qualitative and quantitative results with different values of $Th_b$ on a GMM detected frame from the "office" sequence. $Th_b = 20$ still obtains the optimal enhancement result for the proposed filter.

To sum up, the enhancements by morphological operation vary from case to case and sometimes even deteriorate the foreground contours when handling coarse masks with a high level of false negatives. Fully-connected CRF contributes little to improving highly-textured boundaries and sparsely detected foreground areas. The MRF class, commonly based on posterior probabilities, often relies on local pixel connectivity and cannot enhance heavily broken foreground objects. Local filters such as the bilateral filter and the guided filter lack coherence between distant pixels, resulting in poor performance in textured regions. The edge-preserving property of the original spatial tree filter is not as good as the local filters. Our method considers both global spatiotemporal foreground information and local pixels relations with edge-preserving property, so it can effectively cope with imperfect coarse detection masks. The above qualitative comparison results show that the proposed image filter has widespread applicability for enhancing different kinds of change detection methods. Moreover, the algorithm is also suitable for both indoor and outdoor scenes.

2) Quantitative results

Table 1 shows the average F-measure comparisons of enhanced change detection masks across 8 enhancement algorithms over 10 consecutive frames in each sequence. In each set of experiments (in a row), the best result is in bold face and the second best is underlined. Obviously, the proposed method, i.e., the integrated filtering, outperforms the other seven methods in all 16 cases. The spatial tree filter has an average rank of 2 across all 8 methods. Moreover, the values of F-measure are increased to be higher than 0.9 after enhancement by our method in all cases, regardless of how the coarse results are. On the "traffic" sequence detected by SpkmeansEM, the proposed method even yields an increase in F-measure by almost 0.34, which means a rise of more than 50% comparing to the original detection performance. The results shown in Table 1 are consistent with the visual results demonstrated in Figs. 5-8, and the superiority of the proposed algorithm is therefore verified from both quantitative and qualitative aspects.

3.2. Parameter tuning

1) Parameters tuning during filtering

a) Parameters in foreground extraction and matching

The threshold of connected area $Th_{area} = 50$. In this paper, we only regard a foreground area containing more than $Th_{area} = 50$ white pixels by connected component labeling (4-connectivity) to be the potential foreground region for further processing.

The distance threshold of foreground matching $Th_r = 50$. The distance of the same moving foreground target between the current frame and a history frame (in a very short time) is restricted within a threshold $Th_r$ fixed at 50. Those potential foreground boxes that satisfy this constraint are allowed to construct the spatiotemporal guidance image set.

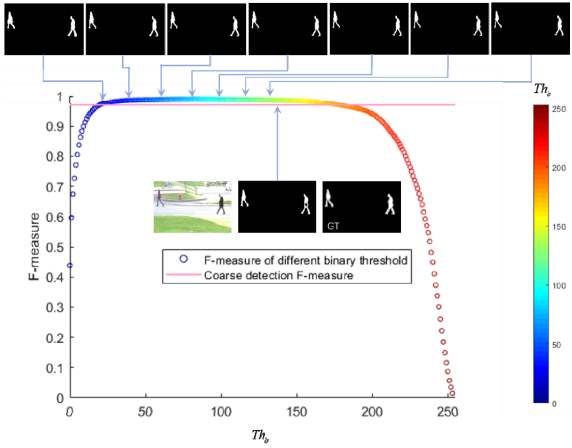

Figure 13: Qualitative and quantitative results with different values of $Th_b$ on a FgSegNet detected frame from the "pedestrians" sequence. We choose $Th_b = 100$ for generating the optimal enhancement result in this case, with all other parameters fixed at recommended numbers.

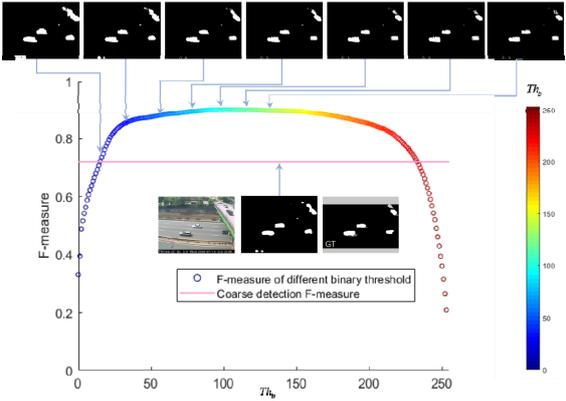

Figure 14: Qualitative and quantitative results with different values of $Th_b$ on a FgSegNet detected frame from the "streetlight" sequence. We choose $Th_b=100$ for generating the optimal enhancement result in this case, with all other parameters fixed at recommended numbers.

b) Parameters for the guided image filtering part

The guided filter part of the integrated filter mainly aims to bring in edge-preserving property. In guided filtering, parameter $r$ controls the size of the moving window for local filtering. Since the spatiotemporal tree filtering has already considered the global information, the value of $r$ should not be too large. Thus, in all experiments $r$ is set to 4. The parameter $\varepsilon$ controls the filtering intensity in the window. And from [27], we know that the parameter $\varepsilon$ controls the extent of edge-preserving property. A large $\varepsilon$ may over blur the filtered image, while a small $\varepsilon$ may weaken the performance. In order to guarantee an adequate local filtering result, we choose $\varepsilon = 0.2^2$ throughout this paper.

c) Parameters of the spatiotemporal tree filtering part

The parameter $k$ decides how many frames are used to construct a guidance set, which is also the spatiotemporal volume for tree filtering. Considering both time efficiency and a common sense that a same object usually appears in a short period of time in a video, the value of $k$ cannot be set too large. But in order to introduce enough spatiotemporal information, $k$ should better be an integer larger than 2. Fig. 9 shows the influence of $k$ on spatiotemporal tree filtering. The experiments are carried out on the "peopleInShade" and the "cubicle" sequences from CDnet, and the "PV" sequence from AVSS2007 dataset [36]. Based on the coarse detection masks obtained by SpkmeansEM, the spatiotemporal tree filtering with $k$ varying from 1 to 4 is utilized to enhance coarse masks and all other parameters are the same. The 1st Row of Fig. 9 is the spatiotemporal filtering results on a foreground region from frame 1972 in the "cubicle" sequence. The upper body of the person is badly detected by SpkmeansEM with a large area of false negatives. With the increase in $k$, the enhancement effect of the upper part of the target improves. Row 2 in Fig. 9 presents the enhancement results of a red moving car from AVSS2007. In the coarse binary mask generated by SpkmeansEM, the windscreen, chassis, and other parts of the car are not fully detected. It can be observed that when $k$ increases, the pixels of the windscreen become whiter, revealing improvement of true positives. Row 3 of Fig. 9 shows enhancement results on a coarse detected region of frame 700 in the "peopleInShade" sequence. In this under-detected foreground region, the enhancement effect of the spatiotemporal filtering is evident and the contour of the person becomes more intact with a bigger $k$. In this paper, $k$ is fixed at 4.

The parameter $\sigma$ adjusts the propagation sensitivity of the global information on MST of a spatiotemporal graph. A larger $\sigma$ makes information on the MST to be more easily transferred, which also means the global information will influence every single pixel more easily. Most change detection algorithms obtain satisfactory results for simple and indoor scenes, but have problems with or even fail on scenes full of background dynamics, illumination changes, and foregrounds with camouflage. For well-detected binary masks, the magnitude of the information transmitted on global MST should not be excessively large to prevent from the foreground leaking to its surrounding background area while the two parts have similar colors in guidance images. On the other hand, for poorly-detected masks, the degree of information propagation on MST needs to be more intense to fully utilize the guidance information in the global scale to support and enhance fragmented foreground parts, and finally to turn the false negatives to true positives. Therefore, we think that the value of $\sigma$ is closely related to the complexity of video scene. For complex scenes, $\sigma$ should better be set to a large number. While for simple scene, $\sigma$ should be small. Via extensive tests and experiments, we set $\sigma \in [5,25]$ for the spatiotemporal tree filtering. For the video sequences appear in section 3.1, "streetlight" and "traffic" are both outdoor traffic scenes, and the former was captured by a shaky camera. Thus, we set $\sigma = 25$ for the two sequences. The "office" sequence is a simple indoor scene, so $\sigma$ is set to 10. The "pedestrians" sequence is a bright outdoor scene simpler than "streetlight" and "traffic" but more complex than "office", so we set $\sigma = 15$.

d) Weights $w_1$ and $w_2$ in integrated filtering

By fixing all parameters mentioned above, we examine the optimal proportions of the local filtering part and the global filtering part to the whole using Receiver Operating Charateristic (ROC) curves. Under the framework of our integrated filter, we vary the binarization parameter $Th_b$ to generate ROC curves with several different ratios between $w_1$ and $w_2$. The ROC tests are based on coarse masks by GMM and SpkmeansEM on sequences of "office", "traffic", "streetlight", and "pedestrians". The horizontal of the ROC figure is the False Positive Rate (FPR) defined as $FP/(FP+TN)$, and the vertical axis of the ROC figure is the True Positive Rate (TPR) that is the same with Recall. In Fig. 10, the case $w_1 = 0$ means the integrated filter degenerates into the guided image filter and $w_1 = 1$ stands for the degeneration into a pure spatiotemporal tree filter. Obviously, both ROC curves of these two extreme cases are below all other curves at different ratios in Fig. 10, which proves that the integrated filter outperforms the filter that only focuses on either global or local aspect. According to the four subplots in Fig. 10 for the comparison of ROC curves under different weights, the integrated filter with $w_1$ set between 0.7 and 0.8 (with $w_2 = 1 - w_1$) will have an optimal enhancement performance because its ROC curve is generally the closest to the upper-left corner of the figure. Thus, in all experiments in this paper we fix $w_1 = 0.8$ and $w_2 = 0.2$.

2) Parameter tuning in binarization

In this section the value of $Th_b$ is separately fixed in two different situations.

We first look at the first situation: the filter is used after a change detection algorithm that tends to get sparse or under-detected foreground contours (which means a large number of false negatives). We obtain the enhanced binary masks by first filtering coarse masks with above recommended parameters and then varying the binarization threshold $Th_b$ from 0 to 255. The final binary masks are contrasted with ground truth in both qualitative and quantitative ways to decide what is the value of $Th_b$ for the optimal enhancement. To form this situation, we use GMM on a frame from "streetlight" and another frame from "office" for generating sparsely detected foreground objects in coarse masks. Fig. 11 demonstrates the changes in F-measure when $Th_b$ increases from 0 to 255 after the GMM mask from the "streetlight" frame, and also shows qualitative results generated by several different values of $Th_b$ (5, 20, 40, 60, 80, 100, and 120). The F-measure curve peaks in the interval of $Th_b \in (10,30)$ and then the value of F-measure slowly decreases. The pink horizontal line in the figure shows the F-measure level of the GMM coarse result. When $Th_b$ is smaller than 10, we can see that the areas of foreground become over-smoothed, introducing a lot of false positives around vehicle boundaries. By comparing the qualitative results given at different binarization thresholds, we find $Th_b = 20$ obtains the optimal enhancement result for the proposed filter in this case. Fig. 12 shows the qualitative and quantitative enhancement evaluations on the GMM mask of a "office" with varying $Th_b$, The qualitative results are generated by several different values of $Th_b$ (5, 20, 40, 60, 80, 100, and 120). In Fig. 12, much foreground leaks from the leg of the person to the surrounding background when $Th_b$ is less than 10. The result at $Th_b = 20$ has the best trade-off between low noise level and high completeness of foreground contours. When the threshold continues to increase, the F-measure and the completeness of the foregrounds both monotonically decrease. According to the analysis above, we fix $Th_b = 20$ for those change detection algorithms (especially the statistical background-foreground models) that are prone to under-smoothed foreground regions.

TABLE 2
COMPARISONS ON THE AVERAGE F-MEASURE VALUES BEFORE AND AFTER THE PROPOSED ENHANCEMENT METHOD. THE EXPERIMENTS ARE CARRIED OUT ON MASKS GENERATED BY THREE DIFFERENT CHANGE DETECTION METHODS ON 14 COMPLETE CDNET SEQUENCES, RESPECTIVELY.

| Categories | Video sequences | Change detection methods | | | | | |
|---|---|---|---|---|---|---|---|
| | | GRBM | | IUTIS_5 | | MD | |
| | | Coarse masks | After our method | Coarse masks | After our method | Coarse masks | After our method |
| Baseline | Highway | 0.9149 | **0.9317** | 0.9856 | **0.9863** | 0.9188 | **0.9842** |
| | Pedestrians | 0.9280 | **0.9494** | **0.9774** | 0.9754 | 0.9370 | **0.9696** |
| CameraJitter | Boulevard | 0.6916 | **0.8301** | 0.7680 | **0.7835** | 0.4423 | **0.4738** |
| | Sidewalk | 0.4194 | **0.4278** | 0.8290 | **0.8987** | 0.3169 | **0.3569** |
| DynamicBackground | Boats | 0.7937 | **0.8293** | 0.7533 | **0.8426** | 0.6730 | **0.8824** |
| | Overpass | 0.8117 | **0.8637** | 0.9303 | **0.9327** | 0.7759 | **0.8463** |
| IntermittentObjectMotion | AbandonedBox | 0.8302 | **0.8948** | 0.9108 | **0.9120** | 0.5950 | **0.6883** |
| | StreetLight | 0.9360 | **0.9742** | **0.9892** | 0.9889 | 0.8049 | **0.9523** |
| Shadow: | Bungalows | 0.9531 | **0.9794** | 0.9817 | **0.9860** | 0.8904 | **0.9741** |
| | BusStation | 0.8972 | **0.9057** | 0.9294 | **0.9305** | 0.8254 | **0.8867** |
| Thermal: | LakeSide | 0.6224 | **0.6669** | 0.6009 | **0.7057** | 0.4200 | **0.5201** |
| | Park | 0.7039 | **0.7493** | 0.7650 | **0.7957** | 0.6076 | **0.8384** |
| Total average | | 0.7918 | **0.8335** | 0.8684 | **0.8948** | 0.6839 | **0.7811** |

For those algorithms that tend to get over-smoothed detection results (e.g., CNN, and FgSegNet). We use a same way to examine the best value for $Th_b$ in application. We obtain the enhanced binary masks by first filtering coarse masks with recommended values for the rest of parameters and then varying the binarization threshold $Th_b$ from 0 to 255 to obtain the final result. The final binary masks are contrasted with ground truth in both qualitative and quantitative ways to decide what is the optimal value of $Th_b$ for enhancement. To form this situation, we apply FgSegNet on a frame from "pedestrian" and another frame from "streetlight" for generating over-smoothed foreground objects in coarse masks. Fig. 13 demonstrates the changes in F-measure when $Th_b$ increases from 0 to 255 after the FgSegNet mask from a frame in "pedestrians", and also shows qualitative results generated by several different values of $Th_b$ (5, 20, 40, 60, 80, 100, and 120). It can be observed that the over-smoothed case is different from the under-detection case. When $Th_b = 20$, both qualitative and quantitative performances are even not comparable to the coarse mask; the enhanced foreground regions are apparently over-smoothed. The F-measure curve shows that the performance peaks around the $Th_b$ value of 100. The qualitative results then show the over-smoothing effect can be significantly alleviated when $Th_b$ is greater than 80. And it can also be observed that the contour of the left people with $Th_b = 100$ in the image is actually more accurate than the case when $Th_b = 80$.

Fig. 14 presents the parameter tuning results of $Th_b$ for enhancing a FgSegNet mask on the "streetlight" sequence. The performance with the interval $Th_b \in (60,150)$ is satisfactory and the F-measure peaks at about $Th_b = 100$. From the qualitative results, only in three cases, the parameter $Th_b$ equals 100, 120, and 140, the two cars on the upper-right corner can be correctly separated. Besides, when $Th_b = 100$, the integrity of the foreground objects is better than the other two cases. According to the above evaluations and analysis, we set $Th_b = 100$ for the change detection methods that tends to obtain over-smoothed foreground regions.

### 3.3. Experiments on some complete sequences

In order to further verify the superiority and applicability of the proposed enhancement method, we apply it to detection masks of three foreground detection algorithms, i.e., GRBM [37], MD [38], IUTIS_5 [39], on 14 complete video sequences in CDnet2012. In CDnet competition, the three methods have medium, low and high ranks,

TABLE 3

COMPARISONS OF THE AVERAGE PROCESSING TIME (IN MILLISECOND) AMONG EIGHT ENHANCEMENT METHODS

| Sequences | Resolution | Change detection methods | Enhancement methods | | | | | | | |
|---|---|---|---|---|---|---|---|---|---|---|
| | | | Morphological operator | FA MRF | Li MRF | Joint bilateral filtering | Guided filtering | CRF | Spatial tree filtering | Our method |
| Streetlight | 320*240 | GMM | 0.34 | 17.00 | 16.30 | 12.32 | 7.07 | 189.90 | 230.61 | 17.73 |
| | | SpcmeansEM | 0.34 | 16.50 | 17.80 | 11.76 | 6.81 | 189.33 | 227.10 | 16.97 |
| | | Vibe | 0.48 | \ | \ | 12.12 | 7.14 | \ | 229.24 | 17.53 |
| | | FgSegNet | 0.36 | \ | \ | 12.07 | 8.47 | 201.22 | 231.80 | 21.19 |
| Pedestrians | 360*240 | GMM | 0.50 | 55.00 | 29.70 | 12.91 | 7.86 | 228.71 | 230.99 | 21.96 |
| | | SpcmeansEM | 0.41 | 18.20 | 27.90 | 13.67 | 8.59 | 242.19 | 232.25 | 21.51 |
| | | Vibe | 0.38 | \ | \ | 13.45 | 7.72 | \ | 236.26 | 22.61 |
| | | FgSegNet | 0.39 | \ | \ | 13.42 | 8.19 | 252.81 | 235.25 | 22.55 |
| Office | 360*240 | GMM | 0.41 | 18.30 | 9.20 | 15.34 | 9.10 | 198.10 | 237.02 | 33.65 |
| | | SpcmeansEM | 0.40 | 37.10 | 18.90 | 13.63 | 8.05 | 189.25 | 237.80 | 36.39 |
| | | Vibe | 0.42 | \ | \ | 25.78 | 7.99 | \ | 238.06 | 33.70 |
| | | FgSegNet | 0.41 | \ | \ | 12.83 | 7.94 | 215.88 | 241.52 | 32.85 |
| Traffic | 320*240 | GMM | 0.86 | 34.20 | 17.30 | 15.42 | 14.68 | 190.87 | 229.05 | 24.87 |
| | | SpcmeansEM | 0.42 | 34.50 | 17.40 | 11.99 | 7.02 | 184.73 | 228.76 | 21.83 |
| | | Vibe | 0.36 | \ | \ | 12.71 | 11.26 | \ | 228.81 | 24.62 |
| | | FgSegNet | 0.89 | \ | \ | 12.14 | 10.39 | 196.37 | 232.11 | 19.50 |

respectively. The 14 sequences, a total of 25525 video frames, are selected from 7 categories in CDnet2012 dataset. The parameters are fixed according to the recommendation in section 3.2. The average F-measure of all sequences are presented in Table 2. The resultant values that are raised by the proposed enhancement method are in bold face. From Table 2, we can see that except two cases in which the performance of coarse detections are already high, the proposed method is effective in enhancing results of the three different change detection algorithms for almost all sequences. The integrated filter seems particularly suitable for improving the masks generated by a barely satisfactory change detection method. For the MD, nearly half of the sequences are improved by more than 10% in F-measure after our integrated filtering. However, the proposed filter has very limited enhancement effect on coarse detection masks that have too many false positives. The reason comes from the big difference between the algorithm-extracted foreground and the true foreground, which causes "chasms" on MST of the spatiotemporal graph, rendering the information aggregation along the paths on the MST ineffective.

3.4. Speed

The computational cost of the proposed enhancement method is analyzed by measuring the average processing times for four CDnet sequences, respectively. Eight enhancement algorithms are compared on coarse masks by four different change detections, respectively. Therefore for each enhancement method there exists 16 cases, all recorded processing times are listed in Table 3. Since the integrated filtering is based on the local spatial foreground area and the spatiotemporal guidance set, the computational cost is closely related with the number of the potential foreground areas and the total area of foreground regions in the spatiotemporal volume. In this experiment, the average processing time of our method ranges between 16 milliseconds and 37 milliseconds. The morphological operation is the fastest algorithm but meanwhile it usually performs worst when comparing with other methods. CRF is a dense global enhancement method; the standard CRF (10 iterations) costs about 200ms. The original spatial tree filter which runs on a complete frame costs about 200ms in average. Although our method does not lead in speed comparing to some other methods, it satisfies real-time requirements.

4. Conclusion

Change detection algorithms face several problems such as detection noise, inaccuracy in foreground contours, insufficient adaptation to a broad range of complex scenes, and vulnerability to changes in the scene. In this paper, we propose an enhancement method based on integrated filtering to improve the coarse detection masks generated by different types of change detection algorithms. The proposed integrated filter combines both advantages of local image filtering and global image filtering; it utilizes local pixel coherences to preserve edges between background and foreground, and uses the self-similarity of the same foreground object in consecutive frames to aggregate global information for filtering in a spatiotemporal domain. The proposed method is essentially a two-scale method that fuses information from the local scale and the global scale, therefore it is efficient and effective to refine the detection mask to get closer to the ground truth.

Experiments show that our method is able to enhance detection results of different types of foreground detection algorithms span from statistical background modeling to deep learning methods. In addition, it has been demonstrated to outperform many other popular enhancement methods from qualitative and quantitative aspects. Currently, despite the guidance can be a colorful image, the input of the integrated filtering is only binary. In the future, we aim to further generalize the enhancement framework to a broader range of applications such as image saliency detection, stereo vision, and multi-category semantic segmentation. We are also interested in introducing advanced target tracking techniques into the potential foreground target matching step in the workflow of the proposed detection enhancement to see whether the performance can be further improved.